\documentclass{tlpHack}
\usepackage{aopmath}
\usepackage{amscd}
\newtheorem{definition}{Definition} 
\newtheorem{example}{Example} 

\long\def\comment#1{}
\def\whiteghost#1{{\setbox1=\hbox{#1}\hbox to
\wd1{\vrule width 0dd depth
\dp1 height \ht1 \hfil}}}

\def\lc{\mathop{\mbox{\tt-\char'134}}} 
\def\rc{\mathop{\mbox{\tt/-}}} 
\def\prop{\mathop{\mbox{\tt::>}}} 
\def\simp{\mathop{\mbox{\tt<:>}}} 
\def\chrprop{\mathop{\mbox{\tt==>}}} 
\def\chrsimp{\mathop{\mbox{\tt<=>}}} 
\def\ttbar{\mbox{\tt|}} 

\def\busteps{\mathop{\mbox{\hbox to 0pt{$^*$\hss}$\nearrow$}}}

\def\tdsteps{\mathop{\mbox{$\searrow$\hbox to 0pt{\hss$^*$}}}}

\def\bang{\mathopen{\mbox{\tt!}}}
\def\parmatch{\mathop{\mbox{\tt\char'44\char'44}}} 
\def\ttcomma{\mathclose{\mbox{\tt,}}}

\def\ttbackslash{\tt\char'134}
\def\true{\mbox{\tt true}}
\def\false{\mbox{\tt false}}

\def\Grammar{\mbox{\it Grammar}}
\def\Context{\mbox{\it Context}}
\def\IC{\mbox{\it IC}}
\def\Discourse{\mbox{\it Discourse}}
\def\Phrases{\mbox{\it Phrases}}
\def\Phrase{\mbox{\it Phrase}}
\def\Constituents{\mbox{\it Constituents}}
\def\Facts{\mbox{\it Facts}}

  \submitted{31 August 2002}
  \revised{26 September 2004, 15 June 2004}
  \accepted{9 August 2004}

\title{CHR Grammars}
\author[H. Christiansen]
{HENNING CHRISTIANSEN\\
Roskilde University, Computer Science Dept.,\\
P.O. Box 260, DK-4000 Roskilde, Denmark\\
\email{henning@ruc.dk}}

\pagerange{\pageref{firstpage}--\pageref{lastpage}}
\volume{\textbf{14} (3):}
\jdate{September 2004}
\pubyear{2004}
\begin{document}

\label{firstpage}
\maketitle

\begin{abstract}
A grammar formalism based upon CHR is proposed analogously to
the way Definite Clause Grammars are defined and implemented
on top of Prolog. These grammars execute as robust bottom-up
parsers with an inherent treatment of ambiguity and a high
flexibility to model various linguistic phenomena. The formalism
extends previous logic programming based grammars with
a form of context-sensitive rules and the possibility to include extra-grammatical
hypotheses in both head and body of grammar rules.
Among the applications are straightforward  implementations of
Assumption Grammars and abduction
under integrity constraints for language analysis.
CHR grammars appear as
a powerful tool for specification and implementation of language
processors and may be proposed
as a new standard for bottom-up grammars in logic programming.
\end{abstract}

\section{Introduction}
Constraint Handling Rules~\cite{fruehwirth-98} (CHR) provide a natural framework
for extending logic programming
with bottom-up evaluation which, together with other qualities of
CHR, makes it interesting  to consider CHR for language processing.
In general, constraint solving techniques have proved to be
important for expressing and solving linguistic problems.

In order to promote and facilitate language processing in CHR, we
propose a standard for a grammar notation built upon CHR,
called CHR Grammars or CHRG for short.
At a first glance, CHRG may be seen as a bottom-up 
counterpart to the well-known Definite Clause Grammars~\cite{PereiraWarren1980} (DCG),
but the CHRG formalism 
includes additional facilities that are not obvious or possible in DCG.
Most notably, the notation supports context-sensitive rules that may consider
arbitrary symbols to the left and right of a sequence be matched.
Counterparts to the different sorts of rules of CHR (propagation, simplification,
and simpagation) are present in CHRG and grammar rules may also refer to
extra-grammatical hypotheses in both head and body of rules.
CHRGs are implemented by a compiler into CHR analogously to the way
DCGs usually are translated into Prolog.
This provides a seamless integration with CHR and Prolog, so that the
high-level notation of CHRG is combined with the sort of tools and
libraries that are relevant for practical applications.

When executed as a parser, a CHRG is  robust of errors and provides an
elegant handling of ambiguity: Rules apply bottom-up as long as possible
and grammar nodes corresponding to the different parses can be read out of the
final constraint store.

The context-sensitive rules provide a high degree of expressiveness both
for simplifying the overall grammar structure and for modeling phenomena
such as long-distance reference and coordination in natural language.
Context-sensitivity can also be used for classifying lexical tokens 
in a way quite similarly to the component called a tagger in language processing systems.

The possibility to apply extra-grammatical constraints in grammar rules
makes it straight\-forward to express abductive language interpretation
with integrity constraints written as CHR rules; no extra meta-level
overhead is necessary.
Facilities from Assumption
Grammars (AG) are included in CHRG in a similar way;
AGs are in many ways similar to abduction but provide
also primitive scoping mechanisms not found in the abductive
approach.

The CHRG system accepts any grammar whose context-free backbone
is without empty-productions and loops and it has no problems with left-recursion
as is the case for DCG. The efficiency is highly dependent on the grammar:
For locally unambiguous grammars (to be defined), execution is linear
and for a general context-free grammar cubic similarly to other general
parsing algorithms.

The CHRG system is implemented in SICStus Prolog and is available on the
Internet~\cite{CHRGwebsite}.

\subsection*{Overview}
The following section~\ref{background-section} provides the background and motivation
of this work and reviews important, related work.
Section~\ref{CHRG-section} describes syntax and semantics of the CHRG notation together
with the principles used for its implementation in CHR; section~\ref{examples-section} shows examples of CHRGs.

The approach to abductive language interpretation is described in section~\ref {abduction-section},
firstly at an abstract level as a general method for transforming abductive language interpretation
into a deductive form which is not tied to a specific grammar formalism.
We then show how the principles can be applied in CHRG in a version for unambiguous
grammars and an extension for ambiguous grammars (some extra machinery is needed
as to avoid cluttering up abducibles for different parses).
Section~\ref{assumption-grammar-section} explains the implementation of Assumption Grammars
in CHRG.
Section~\ref{summary-section} gives a summary and discusses future perspectives.

\vfil\eject
\section{Background and related work}\label{background-section}
Our work can be described as filling out the lower right
corner in the following commutative diagram:

\smallskip
\begin{center}
$\begin{CD}
\mbox{Prolog}@>>>\mbox{CHR}\\
@VVV @VVV\\
\mbox{DCG}@>>>\mbox {\setbox7=\hbox{CHR}\hbox  to \wd7{\it CHR Grammars\hss}}
\end{CD}$
\end{center}
\smallskip\noindent
Definite Clause Grammars~\cite{PereiraWarren1980} (DCGs) have been an integral
component of most Prolog systems for decades and are basically a derivative of
Colmerauer's Metamorphosis Grammars~\cite{Colmerauer1975}
that were designed together with one of the first versions of Prolog.

DCGs are syntactic sugar for Prolog programs which in their now standard implementation
represent strings by means of difference lists. When executed as a parser,
a DCG inherits  Prolog's
top-down strategy with backtracking for checking out different alternatives.
DCGs are very popular as they are very easy to write down and get running, especially
for toy languages and not too complicated fragments of, say, natural language
or programming languages. DCGs put very few restrictions on the context-free
backbone of the grammar, as do most traditional methods for writing parsers;
see, e.g.,~\cite{AhoSethiUllman1986}.  
The main drawbacks of DCGs are
\begin{itemize}
\item
lack of robustness, if the string to be analyzed does
not
conform with the grammar the result is simply failure,
\item backtracking may lead to combinatorial explosions, so a grammar
for a larger application needs to be tuned very carefully with 
cuts and the like to avoid this,
\item lacking ability to handle left-recursive grammars.
\end{itemize}
To compensate partly for this, different authors (not referenced here) have proposed
compiling DCGs into bottom-up parsers by traditional means.

The CHR language~\cite{fruehwirth-98}
was introduced as a
tool for writing constraint solvers in a declarative
way for traditional constraint domains
such as real or integer numbers and finite domains.
CHR has proved to be of more general interest
and is available as extension of, among others,
SICStus  Prolog~\cite{sicstus-manual}.
The CHR web pages~\cite{chronline} contain a growing
collection of applications.
Being of special interest to language processing,
\cite{AbdennadherSchuetz98} have shown that CHR adds
bottom-up evaluation to Prolog and a flexibility
to combine top-down and bottom-up computations;
\cite{AbdChr2000} have taken this a step further showing that
abductive logic
programs can be expressed directly in CHR.

The metaphor given by the diagram above is very precise as we propose a notation
that can be seen as a layer of syntactic sugar over CHR rules that parses
bottom-up. A string is entered as a set of initial constraints and the rules
apply over and over producing more and more syntax nodes from
those already found. In this way we achieve a robustness not found in
DCGs and avoid also the problems with backtracking and left-recursion;
furthermore, this approach gives an inherent and elegant treatment of ambiguity
without backtracking. In our approach, a string is encoded by means
of integer word boundaries as also used in
Datalog grammars~\cite{DTH} and the
classical paper on DCGs~\cite{PereiraWarren1980}.

It is interesting to compare our CHRG  formalism with
Constraint Multiset Grammars~\cite{Marriott94} (CMGs)
that can apply also to multidimensional
languages such as diagrams. The rules of CMG include also
context-conditions which seems capable of expressing the sort
of context conditions included in CHRG.
Meyer~\cite{Meyer2000} has applied CHR for parsing of
CMGs with techniques very similar to ours, however
without considering the compilation of a grammar notation into
CHR. Recent work by~\cite{bottoni-et-al-2001} has proposed to
apply a variant of linear logic for parsing CMGs.

Morawietz~\cite{Morawietz-2000} has implemented
deductive parsing~\cite{ssp95}
in CHR and shown that a specialization of a general bottom-up
parser for context-free rules
leads to propagation rules similar to those produced by
our compiler.
Our proposal for a grammar notation upon CHR
was put forward in~\cite{hc2001} and presented briefly in~\cite{christiansen-iclp02};
the CHRG system has also been presented as~\cite{christiansen03}.
An attempt to characterize the grammar of ancient Egyptian hieroglyph
inscriptions by means of context-sensitive rules in CHRG is given
by~\cite{hnp-2002}.

In~\cite{ChristiansenDahl2002,ChristiansenDahl2003}, we have applied CHR for parsing
with error detection and correction in which we employ CHRs
ability to combine top-down and bottom-up computations,
cf.~\cite{AbdennadherSchuetz98}:
Parsing proceeds bottom-up as described in the present paper
and when symptoms of an error are seen, a top-down sweep for
correcting the string is started, so that the parser may continue.

The notion of constraints, with slightly
different meanings, is often
associated with language processing.
``Constraint grammars'' and ``unification grammars''
are often used
for feature-structure grammars,
and constraint programming techniques have been applied
for the complex constraints that arise in natural
language processing; see, e.g.,~\cite{GazdarMellish1989,Allen1995,Duchier2000}
for introduction and overview.
One approach using CHR for this purpose in HPSG is~\cite{Penn-2000}.

Blache~\cite{blache2000} proposes a formalism with specific
kinds of constraints
for natural language which also seems to fit with an implementation
in CHR. This approach combines constraints on the order in which
things must occur, on which things
imply the presence or absence of other things, etc.
We have not tried to model this in CHRG, but CHRGs contexts
and possible use of arbitrary hypotheses seems to
be suited.
See also~\cite{Duchier99a,Maruyama1990,smfs2000} for similar approaches.

Our approach to abductive language interpretation using CHR Grammars
is based on extension of our previous work~\cite{AbdChr2000} who showed how
an abductive logic program with integrity constraint (but limited use of negation)
can be rewritten as a CHR program;
basically, the idea is to declare abducibles as constraints
and write integrity constraints as CHR rules, and abduction
works so to speak for free without any meta-level interpreter which usually
is associated with abduction.
We are not aware of other approaches
to abductive language interpretation using CHR in this way.

The advantages of abduction for language interpretation
--- as theoretical model or as implementation ---
has been recognized be several authors,
e.g.,~\cite{charniak-mcdermott-85,gabbey-kempson-pitt-94,weighted-abduction,mig-kobe-93}
just to mention a small fraction,
and this
is taken in the present work as an established fact.

A conventional implementation of
DCGs~\cite{PereiraWarren1980} applies a purely
deductive interpretation method, 
synthesizing the meaning of a phrase from
the meanings of its subphrases.
This works well when context is known and
every piece of information to be extracted is expressed in an
explicit way.
Abduction is in favour  for more subtle meanings given, e.g, by linguistic 
implicature,
and when the attention is on context comprehension.
In~\cite{ChristiansenContext99} we have related
Stalnaker's~\cite{stalnaker-98}
view of context comprehension to abductive language interpretation.
One way to achieve abduction with logic grammars is, of course, to interpret
a DCG using an interpreter for abductive logic programs such as~\cite{kakas-et-al-2000};
we have not made any benchmark tests but we expect this to
be far less efficient than what is described in the present paper.
An interesting variation of our method is
to combine the core of our abduction method with DCGs as shown
in example~\ref{dcg-with-abduction-example} below:
The DCG is processed in the usual way but it may refer to abducible predicates
defined as CHR constraints.
An earlier paper~\cite{christiansen-nlulp02}
on our approach to abductive language interpretation
discusses in more detail the relation to other abduction methods.
In~\cite{ChristiansenDahl2004} we have considered how
our CHR versions of abduction
and assumptions~\cite{DahlTarauLi97} with integrity constraints
can be used as an extension to Prolog.

\section{Syntax, semantics, and implementation of CHRG}
\label{CHRG-section}
\subsection{Preliminaries: First-order logic and CHR}\label{CHR-section}
First-order logic is assumed; variables are typically denoted by letters such as $x$,
$y$, $\ldots$ or with capital letters in typewriter font in programming notation;
constants are typically denoted by letters such as $a, b, \ldots$;
notation with a horizontal bar as in $\bar x$ refers to a sequence of variables,
similarly $\bar a, \ldots$ for sequences of constants and
$\bar t, \ldots$ for sequences of
terms.

We give the necessary definitions and properties for Constraint Handling Rules (CHR)
in a slightly simplified form and
refer to a general introduction elsewhere \cite{fruehwirth-98}.

Two disjoint sets of constraint predicates are assumed,
called  {\em defined constraints} (i.e., defined by the current
program) and {\em built-in constraints},
the
latter including ``$=$'',  ``$\neq$'', $\true$, and $\false$.
Atoms of constraint predicates are (with a slight overloading of usage) called
({\em defined} and {\em built-in}) {\em constraints}.
Conjunctions are written by either ``$\land$'' or, in programming notation, a comma.

The following CHR rules\footnote{Our usage is to consider ``CHR'' as a name of
a language rather that a written shorthand for a three-word term, thus ``CHR rule''
is not redundant.} are recognized:
\begin{description}
\item[Progagation rules] $H \chrprop G\ttbar B$,
\item[Simplification rules] $H \chrsimp G\ttbar B$, and
\item[Simpagation rules] $H \backslash H' \chrsimp G\ttbar B$ being an abbreviation
for $H\ttcomma H' \chrprop G\ttbar H\ttcomma B$.
\end{description}
Each $H$ (and $H \backslash H' $) is called the {\em head}\footnote{Our
terminology differs slightly from~\cite{fruehwirth-98} who refers to {\em each} atom to the left
of the arrow as a head.}  of the rule and is
a conjunction of one or more defined constraints indicated by commas,
$G$ the {\em guard} being a conjunction
of built-in atoms, and $B$ the {\em body} being a  conjunction of constraints.
A guard corresponding to $\true$ can be left out together with the vertical bar.

In examples and extensions to the framework we apply occasionally the possibility
in the implemented CHR system of including arbitrary Prolog code
in rule bodies, including those auxiliaries of the CHR system that goes beyond
a declarative semantics as well as the abstract, procedural semantics given below. 
The same goes for the application of so-called deep guards
in which constraints are called in the guard.
In such cases we supply suitable informal descriptions.

A CHR {\em program} is a finite set of rules with its declarative semantics given
as the conjunction of a logical reading of each rule as follows; the built-in
``$=$'' and ``$\neq$'' have their standard syntactic meaning.
Propagation rules and simplification rules in the format above
are taken as abbreviations for the following respective formulas:

\[ \forall \bar x \bigl((\exists\bar y G) \rightarrow (H\rightarrow\exists \bar z B)\bigr) \]

\[ \forall \bar x \bigl((\exists\bar y G) \rightarrow (H\leftrightarrow\exists \bar z B)\bigr) \]
where $\bar x$ refer to the variables in $H$, $\bar y$ to those in $G$ not overlapping with
$\bar x$, and $\bar z$ to
those in $B$ not overlapping with
$\bar x$; for simplicity it is assumed that $\bar y$ and $\bar z$ do not overlap;
see~\cite{fruehwirth-97,fruehwirth-98} for a generalization.
A rule with $\bar z$ empty is said to be {\em range-restricted}.

A {\em state} is defined to be a set of constraints and an {\em initial} state for a {\em query}
$Q$ (being a conjunction of constraints) is $Q$ itself; we do not distinguish between sets and
conjunctions. 
We distinguish a special state referred to as {\em failure} and any derivation step
(below) leading to this state is said to be {\em failed}.

To {\it execute a(n instance of a) body} $C\land E\land N$ where $C$ are defined constraints,
$E$ and $N$ built-in's with predicates ``$=$''  and ``$\neq$'', resp., in
state $S$, consists  of forming the state $(S\cup C\cup N)\sigma$ where
$\sigma$ is a common, most general unifier for $E$.
In addition, any $s\neq t$ with $s$ and $t$ nonunifiable is removed.
However, if no such $\sigma$ exists
or $(S\cup N)\sigma$ contains $t\neq t$ for some term $t$, the execution
fails.
Execution of a body containing $\false$ fails.

For an instance $H \chrprop G\ttbar B$ of a propagation rule, we say that it 
{\em can be applied} in a state $S$ whenever $H\subseteq S$ and
$S\models\exists G$, and to {\em apply} it means to execute $B$ leading to a new state.
When referring to an application of a rule $H \chrprop G\ttbar B$ of the current program,
this refers
to some application of an instance $(H \chrprop G\ttbar B)\sigma$ where
$\sigma$ is a substitution to the variables of $H$ (referred to as $\bar x$ above).
No rule can be applied to a failure state.
Application of simplification rules is defined in a similar way
except that the head constraints are removed from the state before the body is executed.

A {\em derivation} for a query $Q$ with program $P$ is a sequence
of states $Q=S_0, S_1, \ldots, S_n$ where each $S_i$, $0<i\leq n$ is
the result of applying a rule of $P$ to $S_{i-1}$ with  $S_{i-1}\neq S_i$.
A given propagation rule cannot be applied to the same
constraints more than once.
A state in a derivation is {\it final} if it is not failed and no rule can apply, and
in this case the derivation is {\em successful};
a derivation ending with a failure state is said to be {\em failed}.

In practice, CHR programs are executed in a specific left-to-right order which
may or may not restrict the
final result. To define this, we must pay attention to the order in which conjunctions
are written and the textual order of the rules;
the actual computation rule applied in, say, the SICStus Prolog version of
CHR~\cite{sicstus-manual} is quite complicated but the following simplified characterization is a good approximation
that covers most cases.
An {\em LR-derivation}  is one
in which:
\begin{itemize}
\item A state is a sequence of constraints $c_1\ttcomma\ldots\ttcomma c_n$.
\item A built-in constraint is considered (as specified above) only when it
appears as $c_1$   and this takes priority over rule applications.
\item
For all $i$, no rule application involves any
of $c_i\ttcomma\ldots\ttcomma c_n$ if another application of a rule
is possible.
\item Rules are tested for applicability in the textual order in which they occur
in the program.
\item Whenever a rule is applied in a step, requiring constraints $R$ to be removed
from and $A$ (as a sequence given by textual order in rule body) to be added to a state
$S=c_1\ttcomma\ldots\ttcomma c_n$, the new state is
$A \ttcomma S'$
where $S'$ is $S$ with $R$ removed and with the order of the remaining 
constraints preserved.
\end{itemize}
This principle is also referred to as the
{\em LR computation rule} and it implies that there is only
one possible derivation.
The version of CHR that underlies the implemented CHRG system~\cite{CHRGwebsite}
performs LR-derivations.
A derivation without this computation rule is called {\em unrestricted}.

The following correctness properties for
CHR derivations follow from~\cite{fruehwirth-98}:

\begin{proposition}[Soundness]
Let $P$ be a CHR program, $Q$ a ground query, and $F$ a final state in a derivation
for $Q$. Then $P\models Q\leftrightarrow\exists F$ and $P\cup Q\models \exists F$.
\end{proposition}

\vfil\eject
\begin{proposition}[Completeness]
Let $P$ be a CHR program and $Q$ a ground query
which has at least one finite derivation and let $F$ be a conjunction of constraints
so that $P\models Q\leftrightarrow \exists F$.
Then there exists a derivation with final state $F'$ so that
$P\models \exists F'\leftrightarrow \exists F$.
\end{proposition}
The following consequences are relevant
for soundness and completeness of bottom-up parsers written in CHR.
\begin{proposition}
Let $P$ be a CHR program consisting of range-restricted propagation rules only
and let $F$ be a final state for a ground query $Q$.
Then $F$ is the least Herbrand model for $P\cup Q$.
\end{proposition}
In our treatment of abduction we may occasionally arrive at rules that are not
range-restricted so the following refinement is useful:
\begin{proposition}
Let $P$ be a CHR program consisting of propagation rules only
and let $F$ be a final state for a ground query $Q$.
Then there exists a ground instance of $F$ which is a
least Herbrand model for $P\cup Q$.
\end{proposition}
When using CHR for checking integrity constraints we rely on:
\begin{proposition}
Let $P$ be a CHR program with the property that any derivation with $P$ is
finite. We have, then, that $P\cup\exists Q$ for any query $Q$ is consistent if and only if
there is a successful derivation for $Q$ with $P$.
\end{proposition}
Soundness of  disambiguation of grammars by replacing propagation rules by simplification
rules follows from:

\begin{proposition}
Let $P$ be a CHR program consisting of propagation rules,
and $P'$ derived from $P$ by changing
some rules into simplification or simpagation rules, and let $S$ and $S'$ be
final states for a given query with the programs $P$ and $P'$.
Then $S'\subseteq S$.
\end{proposition}

\subsection{Syntax and informal semantics of CHRG}
A {\em CHR Grammar}, or {\em CHRG} for short consists of 
finite sets of  {\em grammar symbols} and {\em constraints}
and a finite set of {\em grammar rules},
each of which may be a {\em propagation (grammar) rule},
a {\em simplification (grammar) rule}, or a {\em simpagation (grammar) rule}.

An {\em attributed grammar symbol}, for short  called a {\em grammar symbol},
is formed as an atom whose predicate symbol is a grammar symbol;
a grammar symbol formed by {\tt token}/1 is called a {\em terminal},
any other grammar symbol a {\em nonterminal}.
Sequences of terminal symbols {\tt token($a_1$), $\ldots$, token($a_n$)}
may also be written {\tt[$a_1$, $\ldots$, $a_n$]}; if ground, such
a sequence is called a  {\em string}.

A few grammar symbols and operators are given a special meaning (made precise later):
\begin{itemize}
\item ``{\tt ...}'' and ``{\tt$i$...$j$}'' with $i<j$ called {\em gaps}\footnote{These gaps provide a superficial resemblance
with Gapping Grammars~\cite{dahl84}, however, in the present version of CHGR
it is not possible to move around the string matched by a gap as
in Gapping Grammars.} supposed
to match sequences of arbitrary length, resp., length $n$ with $i\leq n\leq j$,
\item ``{\tt all}'' referring to the entire input string, which may be useful together
with:
\item ``$\alpha\parmatch\beta$'', called {\em parallel match},
supposed to match strings that are matched by
$\alpha$ as well as $\beta$.
\end{itemize}
When referring to a sequence of grammar symbols, this may involve applications
of the parallel match operator.

A propagation rule is of the form
\[\alpha\;\;\lc\;\;\beta\;\;\rc\;\;\gamma\;\;\;\prop\;\;\; G\;\;\ttbar\;\;\delta\mbox{.}\]
The part of the rule preceding the arrow $\prop$ is called the
{\em head}, $G$ the {\em guard}, and $\delta$ the body;
$\alpha, \beta,\gamma,\delta$ are sequences of grammar symbols
and constraints so that
$\beta$ contains at least one grammar symbol,
and
$\delta$ contains exactly one grammar symbol which is a nonterminal
      (and perhaps constraints);
$\alpha$ ($\gamma$) is called {\em left (right) context} and
$\beta$ the {\em core} of the head; $G$ is a conjunction of built-in constraints
as in CHR and no variable in $G$ can occur in $\delta$.
If left or right context is empty, the corresponding marker is left
out and if $G$ is empty (interpreted as {\tt true}), the vertical bar
is left out.
The convention from DCG is adopted that constraints (i.e., non-grammatical
stuff) in head and body of a rule are enclosed by curly brackets).
Gaps and parallel match are not allowed in rule bodies.

There is a restriction on the use of gaps in the core of a head
so that the core must be {\em bounded} defined in the following way.
This ensures that the core matches a specific interval of word boundaries when applied
(and thus defines meaningful boundaries for the body):
\begin{itemize}
\item The core is {\em bounded} if it is {\em left} and {\em right bounded}. 
\item A sequence $A_1,\ldots,A_n$ is left bounded
(right bounded) if $A_1$ ($A_n$) is not a gap. 
\item A parallel match $A\parmatch B$ is left bounded (right bounded) if at least one of
$A$ and $B$ is left bounded (right bounded). 
\end{itemize}
Furthermore, it is assumed that any variable appearing in body as well as guard
also must occur in the head.
A grammar rule is {\em range-restricted} if any variable in the body appears in the head.

A {\it simplification (grammar) rule} is similar to a propagation rule
except that the arrow is replaced by $\simp$;
a {\it simpagation (grammar) rule}  is similar to a simplification
except that one or more
grammar symbols or constraints in the core of the head
are prefixed by an exclamation
mark ``$\bang$''.
The intended meaning is that head core 
elements under a derivation are removed, except those
prefixed by  ``$\bang$''.
(As the order of the elements in the head of a grammar rule does matter,
we cannot take over the syntax from CHR.)
\begin{example}\label{simple-grammar-example} 
The following source text shows the actual syntax used in the implemented
system. The ``{\tt handler}'' command is a reminiscent from the underlying
CHR system; grammar symbols are declared by
the \verb!grammar_symbols! construct as shown; constraints to be used
in grammar rules are declared as in CHR which will be shown in subsequent examples.
The final command has no effect in the present example, but it adds
extra rules needed for the extensions of CHRG described in
sections~\ref{abduction-section} and~\ref{assumption-grammar-section}.
\begin{verbatim}
    handler my_grammar.
    grammar_symbols np/0, verb/0, sentence/0.
    np, verb, np ::> sentence.
    [peter] ::> np.
    [mary] ::> np.
    [likes] ::> verb.
    end_of_CHRG_source.
\end{verbatim}
When the string ``peter likes mary'' is entered word by word, the words are recognized
as a respectively   {\tt np}, {\tt verb}, and {\tt np} in that order, and then the rule
for {\tt sentence} can apply. Since this grammar consists of propagation rules,
the lexical {\tt token}s as well as the  {\tt np}s and {\tt verb} are not consumed.
If we added a rule, say {\tt np, [likes] ::> sentence1}, a {\tt sentence} as well as
a {\tt sentence1} would be recognized. If all rules were changed into
simplification rules, i.e., replacing {\tt ::>} by {\tt <:>}, only one of
 {\tt sentence} and {\tt sentence1} would be recognized.
\end{example}
Left and right contexts of a rule
may include ``disjunctions'' denoted by semicolon
of different alternatives, and this is considered syntactic sugar for
the set of different rules, taking one alternative for the left and one for the right.
\begin{example} 
The rule
\begin{verbatim}
    (a ; b) -\ c /- (d ; e) ::> f
\end{verbatim}
is an abbreviation for the following four rules: 
\begin{verbatim}
    a -\ c /- d ::> f
    b -\ c /- d ::> f
    a -\ c /- e ::> f
    b -\ c /- e ::> f
\end{verbatim}
\end{example} 
The implemented version of CHRG allows control structures in the body (conditionals
and Prolog-style disjunctions) and arbitrary Prolog goals inside
{\tt\{$\cdots$\}} as well as bodies with no grammar symbols;
for the reason of simplicity, we ignore these options in this presentation.

\subsection{Bottom-up derivations as semantics and the relation to 
top-down syntax derivations}\label{translate-CHRG-to-CHR-subsection}
In order to capture the whole CHRG formalism, a semantic definition needs
to be based on bottom-up derivations and the simplest way to achieve this
is by a translation of CHRG into CHR.
For comparison with traditional grammar formalisms, we provide also
a definition of top-down derivations that characterize a subclass of GHRGs.

For each grammar symbol $N$ of arity $n$, we assume a corresponding constraint
also denoted by $N$ of arity $n+2$ called an {\em indexed grammar symbol},
with the extra two arguments referred to as phrase (or word) {\em boundaries}.

For a grammar symbol $S=N${\tt($\bar a$)},
the notation  $S^{n_0,n_1}$ refers to the
indexed grammar symbol
$N${\tt($n_0$,$n_1$,$\bar a$)} with integers $n_0<n_1$;
in case of a terminal, $n_0+1=n_1$ is assumed.
For any sequence $\sigma$ of grammar symbols
$S_1,\ldots,S_k$ and increasing integers $n_0, n_1,\ldots,n_k$,
we let $\sigma^{n_0,n_k}$ refer to the set
$\{S_1^{n_{0},n_1}, \ldots,S_k^{n_{k-1},n_k}\}$
with the existence of $n_1,\ldots,n_{k-1}$ understood.
For the parallel match operator, we define $(\alpha\parmatch\beta)^{n,m}
= \alpha^{n,m}\parmatch\beta^{n,m}$.
This notation is extended so that for a sequence of grammar symbols and
constraints, we remove all constraints from the sequence, put
indexes on the remaining grammar symbols,
and add again the constraints to the  sequence in their original position.

Gaps are removed from rule heads under this translation but give rise
to inequations to be added
to the guard of the resulting CHR rule; we do not formalize this here but illustrate
the principle in example~\ref{translate-gaps-and-parr-match-example} below.

The translation of rules from CHRG into CHR adds two extra variables to
each grammar symbol and we use a notation analogous to the above
to indicate this. So for a sequence $\sigma$ of grammar symbols
$S_1,\ldots,S_k$ and variables $x_0, x_1,\ldots,x_k$,
we let $\sigma^{x_0,x_k}$ refer to the set
$\{S_1^{x_{0},x_1}, \ldots,S_k^{x_{k-1},x_k}\}$
with the existence of $x_1,\ldots,x_{k-1}$ understood.
The notation is extended to sequences of grammar symbols and constraints
as above so that constraints are unaffected.

The translation of a CHRG $G$ into CHR is denoted $C(G)$ and consists
of the translation $C(R)$ of each rule $R\in G$.
For propagation and simplification rules we have
\[C(\alpha\lc\beta\rc\gamma\prop G\ttbar\delta) =
(\alpha^{x_0,x_1}\ttcomma\beta^{x_1,x_2}\ttcomma\gamma^{x_2,x_3}\chrprop G\ttbar\delta^{x_1,x_2})\mbox{,}\]
\[C(\alpha\lc\beta\rc\gamma\simp G\ttbar\delta) =
(\alpha^{x_0,x_1}\ttcomma\gamma^{x_2,x_3}\backslash\beta^{x_1,x_2}\chrsimp G\ttbar\delta^{x_1,x_2})\mbox{.}\]
Simpagation grammar rules are translated similarly to simplifications except
that those elements of $\beta^{x_1,x_2}$ that 
were preceded by ``$\bang$''
in the original grammar rule
are moved to the left of the backslash.

Notice that a grammar rule $R$ is range-restricted if and only if
the CHR rule $C(R)$ is range-restricted.
\begin{example}
The rule in following source text:
\begin{verbatim}
     constraints h/1.
     grammar_symbols a/0, b/1, d/1, e/2.
     a -\ b(X), [c], {h(Y)} /- d(Y) ::> e(X,Y).
\end{verbatim}
is translated into this CHR rule:
\begin{verbatim}
     a(N0,N1), b(N1,N2,X), token(N2,N3,c), h(Y), d(N3,N4,Y)
       ==> e(N1,N3,X,Y).
\end{verbatim}
\end{example}
\begin{example}\label{translate-gaps-and-parr-match-example}
The translation of gaps and parallel matching into CHR is illustrated for
the following CHRG rules.
\begin{verbatim}
     a, ..., b /- ..., c(X) <:> d(X).
     a$$b ::> e.
\end{verbatim}
They are translated into the following CHR rules:
\begin{verbatim}
     c(N5,_,X) \ a(N1,N2),b(N3,N4) <=> N2=<N3, N4=<N5 | d(N1,N4,X)
     a(N1,N2),b(N1,N2)==>e(N1,N2)
\end{verbatim}
The gap in the context part of the first rule is used in order to make a
``long-distance reference'' to {\tt c}.
\end{example}
Notice that a gap in the head of core of a simplification rule does not imply
the removal of any grammar symbols recognized in the substring spanned
by the particular ``instance'' of the gap.

A {\it (bottom-up) parsing derivation} for a string $\sigma$ with a CHRG $G$ is
a derivation with the CHR program $C(P)$ for the query $\sigma^{0,n}$ where
$n$ is the length of $\sigma$. 
An interesting class of parsing derivations are those that apply
an LR computation rule as in the implemented CHRG system and
for which we describe some optimizations below.

\begin{definition}\label{loop-free-def}
A {\em single-production} is a grammar rule with singleton grammar symbols
in head core and in body. A grammar is {\em loop-free} if there is no chain of
single productions
\[\cdots g_1(\ldots)\cdots\mbox{\tt >>>}\cdots g_2(\ldots)\cdots, \quad
\ldots,\quad
\cdots g_{n-1}(\ldots)\cdots\mbox{\tt >>>}\cdots g_n(\ldots)\cdots\mbox{,}\]
with $g_1=g_n$; here each occurrence of ``{\tt >>>}'' may stand for any
of ``$\simp$'' or ``$\prop$''.
\end{definition}
In order get rid of termination problems once and for all, any CHRG is assumed
to be loop-free.\footnote{It is possible to weaken this definition slightly:
Some chains of single-productions can be allowed provided their arguments
plus non-grammatical hypotheses do not grow in an application of the rule.
As we have assumed a set-based semantics for CHR (as opposed to multi-sets),
we could allow even {\tt p(X)::>p(X)} but not  {\tt p(X)::>p(f(X))}
or {\tt p(X),\{h(Y)\}::>p(X),\{h(f(Y))\}}.}

We notice without proof the following obvious properties.
\begin{proposition}\label{obvious-grammar-props}
\begin{enumerate}
\item Any parsing derivation is finite (as we assume all grammars to be loop-free).
\item Any state in a parsing derivation with a range-restricted grammar
is ground.
\item The final state in an LR parsing derivation for a given string
is unique (up to renaming of
existentially quantified variables that may occur for non-range-restricted grammars).
\item The final state in a parsing derivation with a propagation rule grammar
is unique (up to renaming $\ldots$); thus LR-derivations are
complete for propagation rule grammars.
\item Completeness of LR-derivations does not necessarily hold for a grammar with
simplification or simpagation rules.
\item\label{context-part-prop-part} Let $G$ be a propagation rule grammar without context parts,
and $G'$ be derived from $G$ by adding to some rules context parts and changing
some rules into simplification or simpagation rules, and let $S$ and $S'$ be
final states for a given string with the grammars $G$ and $G'$.
Then $S'\subseteq S$.
This holds also when we restrict to LR-derivation for $G'$ or for both $G$ and $G'$
\end{enumerate}
\end{proposition}
In order to discuss ambiguity, we define syntax trees but
we do not intend that an implementation should generate trees.
\begin{definition}
Let CHRG $G$ and input string $\sigma$ be given. The
set of {\em syntax trees} over $\sigma$  is defined as follows.
\begin{itemize}
\item Any $t=$ {\tt token($a$,$n$,$n+1$)} in $\sigma$ is a syntax tree with top node
$t$.
\item Whenever a rule instance $\alpha\lc\beta\rc\gamma\mbox{\tt >>>}G\mid\delta$,
``{\tt >>>}'' being one of ``$\prop$'' or ``$\simp$'',
is applied in a derivation and $T_1$, $\ldots$,
$T_n$ are trees whose top nodes are
the grammar symbols in $\beta$, then
\[
\mbox{\vbox{\hbox to 3cm{\hfil$\delta$\hfil}
                       \hbox to 3cm{\tt\qquad/\hfil$\cdots$\hfil\char'134\qquad}
                      \hbox to 3cm{$T_1$\hfil$\cdot$\hfil$\cdot$\hfil$\cdot$\hfil$T_n$}}}
\]
is a syntax tree with top node $\delta$.
\end{itemize}
A syntax tree whose top node does not occur in the final state (i.e., it has been
consumed by a propagation or simpagation rule) is called a {\em hidden syntax tree}
and similarly for the node itself.
The set of {\em LR syntax trees} is defined in a similar way, considering only
instances applied in the LR-derivation from $\sigma$ with $G$.
The notions of subtree and proper subtree are defined in the usual way.
\end{definition}
The relevant notion of unambiguity in the context of CHRG is called local
unambiguity and is a stronger property than the usual notion of unambiguity
for context-free grammars.
CHRG works bottom-up with no sort of top-down guidance so even with an unambiguous
grammar (in traditional sense), it may be the case that some subtree becomes part of two different,
larger trees (but only one of these contribute to a tree for the entire string).
\begin{definition}\label{def-unambiguity}
Consider a CHRG $G$ and a derivation for string $\sigma$
and let $\bf T$ be a set of syntax trees with set of top nodes $\bf N$.
The set $\bf T$ (and $\bf N$) is said to be {\em unambiguous} 
whenever,
for any two grammar symbols {\tt p($i$,$j$,$\cdots$)},
{\tt q($k$,$\ell$,$\cdots$)} $\in\bf N$ it holds
that
\begin{itemize}
         \item  if $i\leq k < j \leq \ell$, then $i=k$ and $j=\ell$, and

         \item  if $i\leq k\leq \ell \leq j$, then
	{\tt q($k$,$\ell$,$\cdots$)} is top node of a subtree
	of {\tt p($i$,$j$,$\cdots$)} or the other way round [the last case
	requires single productions in the grammar and
	$\langle i,j\rangle=\langle k,\ell\rangle$].
\end{itemize}
If, furthermore no new syntax tree of the derivation can be added to $\bf T$
without destroying unambiguity, we say that  $\bf T$  and $\bf N$ are {\em maximal}.
A CHRG $G$ is {\em locally unambiguous} if the set of syntax trees in the derivation from
any input string is unambiguous, and  {\em locally LR-unambiguous}
if the set of syntax tree in the LR-derivation from
any input string is unambiguous.
\end{definition}
Maximal unambiguous sets for a given parsing derivation
may overlap, and each such set corresponds
to one possible way of parsing the string.
As we will see later, when doing abduction with ambiguous grammars, it is possible
to extend a grammar so that the different unambiguous sets are kept apart
by means of indexes.

Although CHRG provides an elegant handling of ambiguous grammars, it may 
be relevant to aim at unambiguity, e.g., for efficiency or to avoid mixing up
extragrammatical constraints for different parses.
One obvious way to achieve this is given by the
following which is easy to prove.
\begin{proposition}\label{prop-simplification-grammars-locally-unambiguous}
A simplification rule CHRG is locally LR-unambiguous.
\end{proposition}
Although we have no theoretical result, it seems reasonable to believe that the
local unambiguity of CHRGs is undecidable as is unambiguity for context-free grammars.
If unambiguity is required this can be guaranteed by
proposition~\ref{prop-simplification-grammars-locally-unambiguous} or perhaps 
using a combination of different sorts of rules, in which case the property needs to
be verified.

It should be noticed, that the definition of unambiguous sets does not
take into account left and right context parts of grammar rules.
A rule that produces a node belonging to one unambiguous set may very likely do
so by referring to contextual nodes belonging to other sets. This may be
considered a bug or a feature but it seems to be the only solution that fits with
our general implementation principle.

To compare with traditional grammar formalisms having their meaning
defined by top-down derivations we consider definite clause grammars;
to simplify the comparison, we make a restriction on how variables can
be used.

\begin{definition}
A {\em definite clause grammar (DCG)} $D$ consists of rules of the form
\[A\mbox{\tt-->}B_1\ttcomma\ldots\ttcomma B_n\ttcomma\mbox{\tt\{}G\mbox{\}}\]
where $A$ is a nonterminal, $B_1$, $\ldots$, $B_n$ are grammar symbols, and $G$ a conjunction
of built-in's so that any variable in $A$ and $G$ occurs in some $B_i$.
A DCG is assumed to be loop-free and without single productions (defined in the usual way).

For any ground sequence of grammar symbols $\alpha A\beta$ ($A$ a single grammar symbol),
define the relation $\alpha A\beta\Rightarrow\alpha B_1\ldots B_2\beta$ whenever
there is a rule in $D$ with a ground instance $A\mbox{\tt-->}B_1,\ldots, B_n, \mbox{\tt\{}G\mbox{\}}$
with $G$ satisfied.
The reflexive, transitive closure of $\Rightarrow$ is denoted
$\mathop{\Rightarrow^*}$.
\end{definition}

\vfil\eject
\begin{proposition}\label{compare-with-dcg-prop}
Let $D$ be a DCG and $C$ the CHRG that for each rule in $D$ of the form indicated above contains
\[B_1\ttcomma\ldots\ttcomma B_n\prop G\ttbar A\mbox{.}\]
For ground grammar symbol $A$ and terminal string $\alpha$, the following
statements are equivalent:
\begin{itemize}
\item
$A\mathop{\Rightarrow^*}\alpha$ using the rules of $D$,
\item $A$ is contained in the final state in any parsing derivation for $\alpha$ using rules of $C$.
\end{itemize}
\end{proposition}
The proof is easily made by induction over the length of the derivations.
Combining this with proposition~\ref{obvious-grammar-props}, part~\ref{context-part-prop-part},
we see that a CHRG with context parts corresponds to a DCG with context-sensitive
restrictions on the derivation relation (that are not easily formalized in the setting of DCG).

Finally, notice that CHRG do not provide empty productions.
These, however, are easily mimicked by inserting for each DCG rule
$A\mbox{\tt-->[]}$  grammar symbols $A(0,0)$, $A(1,1)$, $\ldots$ into the initial constraint store.

\subsection{A compile-on-consult implementation}\label{implementation-section}
We describe here very briefly the principles used for the implementation
of CHRG in SICStus Prolog~\cite{sicstus-manual} and describe some additional
features of the implemented system not already covered;
all facilities are described at the online Users Guide to CHRGs available
at~\cite{CHRGwebsite}.

Similarly to DCG and CHR, CHRG is implemented by changing Prolog's reader
so that the terms read are translated into another form before given
to the Prolog compiler (or interpreter).
SICStus Prolog includes a so-called hook predicate called
{\tt term\_ex}\-{\tt pansion}
that can be extended by the user and which is called automatically by the
Prolog reader for each term read from a source file.
The  \verb!term_expansion! clauses defining the CHRG syntax must
work together with those already defined by CHR. The general structure of
the CHRG implementation is illustrated by the following fragment
that treats the \verb!grammar_symbols! declaration:
\begin{description}
\item[{\tt term\_expansion( (grammar\_symbols G), T):-}]\ \\
{\it add 2 to arities of gr.\  sym.\ spec's {\tt G} and add {\tt token}/3 and
a few more to form {\tt C}}{\tt,}\\
{\tt term\_expansion((constraints C), T).}
\end{description}
Similar rules catch terms formed by the operators {\tt <:>} and {\tt ::>},
translate them into CHR rules as described in
section~\ref{translate-CHRG-to-CHR-subsection} above and let the CHR system
translate them further into Prolog rules.\footnote{It is not possible to compile
CHR into ordinary Prolog clauses and the SICStus Prolog implementation
of CHR is based on the low-level library of Attributes Variables.}

The CHRG notation includes counterparts to CHR's {\tt pragma}s and
rule names (in CHR using an {\tt@} operator), but since it
is not possible for override the {\tt term\_ex}\-{\tt pansion}
 clauses given by CHR,
it has been necessary to rename these operators in the CHRG syntax,
{\tt gpragma} and {\tt @@}.

Notice that this sort of implementation makes it possible to mix freely the
rule formats of Prolog, CHR and CHRG, and DCG for that matter.

Finally, the CHRG notation includes a {\tt where} notation which can be applied to
rules of Prolog and CHR as well. We describe it by an example:
\begin{verbatim}
    a(A) -\ B /- ..., q(X,Y)  ::>  {C}, funny_sentence(A,Z)
    where A = ugly(st(r,u,c(t,u,r(e)))),
          B = (np, verb, np),
          C = (append(X,Y,Z), write(Z))
\end{verbatim}
The meaning is that any occurrence in the rule of {\tt A}, {\tt B}, and {\tt C}
is replaced by the indicated term.
The implementation is very simple and one might wonder why this syntax
is not standard in Prolog systems:
\begin{verbatim}
    term_expansion((Rule where Goal), Result):-
        (Goal -> term_expansion(Rule, Result)
         ; write('Error: where-clause failed: <rule> where '),
           write(Goal),nl,write('Compilation stopped'), abort).
\end{verbatim}
The CHRG system includes a number of options of which
the most important is an optimization in the compilation of grammar rules,
so that all but leftmost symbols of core and possible right context
are marked by {\tt passive} {\tt pragma}s;
see the section on CHR of~\cite{sicstus-manual} for a detailed explanation
of these concepts.
For example, with this option the rule 
{\tt np, verb, np ::> sentence} gets compiled into
\begin{verbatim}
    np(X0,X1)#A, verb(X1,X2)#B, np(X2,X3) ==> sentence(X0,X3)
      pragma passive(A), passive(B).
\end{verbatim}
This has significant influence on the efficiency that
we analyze in detail in section~\ref{complexity-subsection} below.
Operationally, the principle means that this rule is not checked for applicability
at the moment when a new {\tt verb} constraint is created as is the case
if no {\tt pragma} {\tt passive} stuff were added.
And, as the system performs LR-derivation, this check for applicability would anyhow fail.
For the {\tt np}s it means that when a new {\tt np} is created, the system does not check if it might
be followed (qua the word boundary arguments) by {\tt verb}, {\tt np};
it is only checked if the new {\tt np} happens to follow some existing {\tt np}, {\tt verb}
sequence.
It can be shown that the semantics is not changed
for propagation rule grammars with only right contexts.
When left and right context or simplification or simpagation rules are used, there are subtle cases where a rule is not applied although it intuitively should be applied.
When this optimization is used for a grammar of simplification rules only,
the constraint store is used effectively as a parsing
stack
in quite the same way as in a traditional LR$(k)$ parser.

For parsing a specific string, the system includes an auxiliary predicate {\tt parse}
that
converts a list of constants to a sequence of calls to {\tt token} constraints.
This predicate may (as an option that can be switched on and off) display
the word boundaries which makes it easy to compare input and result.
Assuming the grammar of example~\ref{simple-grammar-example} above,
we have the following dialogue.
\begin{verbatim}
    ?- parse([peter,likes,mary]).
    <0> peter <1> likes <2> mary <3> 
    np(0,1),
    verb(1,2),
    np(2,3),
    sentence(0,3),
    token(0,1,peter),
    token(1,2,likes),
    token(2,3,mary) ?
\end{verbatim}
This grammar consists of propagation rules; if all are changed into simplification
rules, only {\tt sentence(0,3)} appears as answer.

\subsection{Time complexity}\label{complexity-subsection}
An apparent advantage of CHRG as compared with DCG is
that we avoid the combinatorial explosions that may arise
under backtracking in case a wrong choice of rule is made in
beginning of the string to be analyzed.

Here we give theoretical measures for the running time of CHRGs,
more precisely the CHR rules that are produced by their compilation,
and discuss the behaviour of the implemented system. 

For simplicity, we do not consider context parts or the use of extra-grammatical
constraints.
Without loss of generality, we consider only rules with one or two grammar symbols
in the head.
The CHR rules to consider are, thus, of one of
the following forms,
possibly with $\chrsimp$ instead of $\chrprop$.
\begin{enumerate}
     \item  $A(i,j,\bar t_1)\chrprop B(i,j,\bar t_2)$

     \item  $A(i,j,\bar t_1),B(j,k,\bar t_2)\chrprop C(i,k,\bar t_3)$
\end{enumerate}
We refer  to the so-called meta-complexity theorems
of~\cite{mcallester2000,gm2001,gm2002} for
bottom-up evaluation of logic programs including
deletion. CHR rules, such as those we use,
with one constraint in the body are covered by this scheme.
The main theorem of~\cite{gm2001} gives that time complexity
for reaching a final state
is of order ${\cal O}(n+p)$ where $n$ is number of constraints in an
initial
state and $p$ the number of prefix firings that have appeared in some
state in the derivation.
The number $n$ is the length of the string in our case.
Estimating $p$ is more difficult:
For each rule of type 1 (above), we count the number of occurrences
of $A(i,j,\bar t_1)$ that have occurred in a state; summing for all type 1 rules,
we can limit the contribution by size of grammar times total number of
grammar symbols that have occurred in the derivation.
For each rule of type 2, the prefix firings are of two kinds,

\vfil\eject
\begin{itemize}
\item occurrences of
$A(i,j,\bar t_1)$ (that can be estimated as for type 1), and
\item
occurrences in any state of a pattern matching the entire head
$A(i,j,\bar t_1)$, $B(j,k,\bar t_2)$.
\end{itemize}
The dominant contribution is the last one for type 2 rules,
i.e., for each rule of type 2 and each $C(i,k,\bar t_3)$ occurring in a state,
the possible ways the interval $[i,k]$ can be split up into $[i,j]$ and $[j,k]$
so that some $A(i,j,\bar t_1),B(j,k,\bar t_2)$ have appeared at the same time in
the state during the derivation.

We continue the analysis for two special cases.
\begin{itemize}
\item {\bf Locally unambiguous grammars:}
Each $C(i,k,\bar t_3)$ in some state is created exactly once from a specific
$A(i,j,\bar t_1),B(j,k,\bar t_2)$ combination. Thus the overall time complexity
is proportional to the total number of grammar symbols that have appeared
in the derivation, and we argue that it is of order ${\cal O}(n)$ for a
locally unambiguous grammar: 
Worst case is a binary
     branching everywhere, so a syntax tree over a string of
     length $n$ has $n$ nodes in its deepest layer,
     $\lfloor n/2\rfloor$ in the second
     deepest layer, $\lfloor n/4\rfloor$ in the next one and so.
Summing up, we get at most $2n-1$ tree nodes.

\item {\bf Arbitrary grammars without attributes:}
First of all, let us estimate the maximum number of nodes.
There are ${\cal O}(n^2)$ different substrings of the input string, each of which can represent
up to $g$ different nodes where $g$ is the number of different grammar symbols in the
vocabulary; this is constant, so number of different nodes is ${\cal O}(n^2)$.

Each such node $C(i,k)$ spans  over an interval $[i,k]$,
and the maximum number of ways it can be split up into two subintervals
by some $j$, $i<j<k$,
possibly representing $A(i,j),B(j,k)$, is $k-i-1$.
This adds another factor $n$, so we end up with a total time complexity
of ${\cal O}(n^3)$.
\end{itemize}
The general cubic complexity for context-free grammars
is similar to that of classical
algorithms such as Early and Cocke-Younger-Kasami.
Its interesting to notice that parsing is linear for locally unambiguous grammars
despite the very naive parsing algorithm which simply applies rules
over and over as long as possible.

It is straightforward to show that the results also hold for grammars with context
parts. So if a grammar is made locally unambiguous by a combination of
simplification rules and context parts, it runs in linear time;
the presence of attributes does not affect this.

When attributes are added in the general case, we can have much worse
than cubic complexity as it appears in the following example:

\begin{example}
Consider the grammar
\begin{itemize}
     \item[]{\tt\ [a]::>a(0)\ \ \
     a(T1),a(T2)::>a(t(T1,T2))}
\end{itemize}
For each pair of $i,j$ marking a substring of the input string, there
will be as many different nodes as there are binary trees
with a frontier of $j-i+1$ nodes. It appears that each node is constructed
in a unique way, but the total number of nodes
is given by a terrible combinatorial expression 
far beyond $n^3$.
\end{example}
How do these results compare with practice?
First of all, the optimization in
section~\ref{implementation-section} adding {\tt passive}
{\tt pragma}s to all but rightmost symbols is necessary in order
to achieve an execution as the one assumed in the 
theorem of~\cite{gm2001}.
Secondly, the method behind the implementation of CHR
that we have used (based on attributed variables), as described by~\cite{HolzbaurFruehwirth1999},
indicates that word boundaries should be uninstantiated Prolog variables
to achieve full efficiency and not integers as we have used.

Experiments with Prolog variables for boundaries confirm
these results but even with integer indexes, CHRGs without
too much local ambiguity
execute
equally fast for strings up to  several hundreds of tokens.

Unfortunately, CHR does not construct explicit prefix-firings
during execution, which means that only grammars with at most two grammar symbols
show the expected running times.

It is possible to have the CHRG compiler reduce the size of heads to at most two,
but a general improvement of CHR
so that it incrementally builds prefix firings would solve the problem.
In practice, however, grammars with heads
with up to three or four symbols may run almost linearly provided the
{\tt passive}
{\tt pragma} optimization is used and local ambiguity is limited.

\section{Examples in plain CHRG}\label{examples-section}

\subsection{Disambiguation with simplification and context parts}
It is often the case that an unambiguous grammar, e.g., a context-free
grammar for a programming language, can be written in a much simpler
form as an ambiguous grammar with additional ``disambiguation principles''
specified outside the grammar formalism; see, e.g.,~\cite{AhoSethiUllman1986}.

As we have noticed already, simplification rule grammars are unambiguous
and by means of context parts, we can direct the derivations as to respect
the priorities we have in mind.

\begin{example}
The following simplification rule CHRG is based on a simple and 
highly ambiguous
grammar for arithmetic expressions with addition, multiplication, and
exponentiation. Right contexts have been added which
provides  a conventional operator precedence.
\begin{verbatim}
    e, [+], e /- (['+'];[')'];[eof]) <:> e.
    e, [*], e /- ([*];[+];[')'];[eof]) <:> e.
    e, [^], e /- [X] <:> X \= ^ | e.
    ['('], e, [')'] <:> e.
    [N] <:> integer(N) | e.
\end{verbatim}
\end{example}
In general, both left and right contexts are relevant, and for
natural language application, it may be relevant to disambiguate  some
portions of the grammar in this way but keeping, say, possibilities
of ambiguity at the sentence structure level.

Natural language processing often involves a phase called
tagging in which the different words are classified 
before the ``real'' parsing process takes place.
Tagging is often performed by means of context sensitive rules
that take into account what is immediately to the left and to the right of
the given word~\cite{Brill95}.
Such rules can be expressed in quite natural way in CHRG using
context parts.

\begin{example}\label{tagger-example}
We consider a languages including sentences such
as ``Peter and Paul like Martha and Eve''. The following
rules classify the {\tt name}s as {\tt subject} or {\tt object}
according to their position
relative to the {\tt verb}.
\begin{verbatim}
    name(A) /- verb(_) <:> subject(A).
    name(A), [and], subject(B) <:> subject(A+B).
    verb(_) -\ name(A) <:> object(A).
    object(A), [and], name(B) <:> object(A+B).
\end{verbatim}
\end{example}

\subsection{Long-distance reference in natural language parsing}
Context parts can also be used as a way to access attributes of
grammar symbols at a certain distance. This is relevant
in natural language when a part of a sentence is left out when this part is understood
to be identical to the matching part of a neighbouring sentence.

\begin{example}
Let us extend the language of example~\ref{tagger-example}
with coordination as in ``Peter and Paul likes and Mary hates Martha and Eve'';
The first sentence is incomplete but is understood to borrow its
subject from the second sentence. This can be expressed as follows.
\begin{verbatim}
      subject(A), verb(V), object(B) ::> sentence(s(A,V,B)).
      subject(A), verb(V) /- [and], sentence(s(_,_,B))
           ::> sentence(s(A,V,B)).
\end{verbatim}
For the sample sentence above, the final constraint store contains
{\tt sentence} nonterminals with attributes
{\tt s(peter+paul,like,martha+eve)} and {\tt s(mary,hate, martha+eve)}.
These rules work also in the case when three or more sentences share
a common object.
For analyzing texts consisting of a single sentence,
a rule with a gap could have been used instead:
\begin{verbatim}
      subject(A), verb(V), /- [and], ..., object(B)
           ::> sentence(s(A,V,B))
\end{verbatim}
\end{example}

\subsection{Post-parsing processing in CHRG}
In an application program using CHRG for text analysis it may be relevant
to make some formatting of the constraint store produced by the parser.
As we have noticed, parsing with an ambiguous propagation rule grammar may
result in a large number of nodes, most of them not relevant for
the further processing (but necessary to guide parsing).
It may be the case that we do not want to reduce ambiguity
in the grammar, so some elaboration of the constraint store
needs to take place following parsing.
Part of such post-parsing processing can in fact be specified conveniently
in CHRG.

\begin{example}
Assume we are scanning a text for noun phrases  ({\tt np}s) by means of
a highly ambiguous grammar with a detailed description of sentence structure
as a way to obtain a high degree of precision in the parser.
When the parser has finished its job, we are only interested in noun phrases
and let us suppose that only maximal noun phrases are of interest,
maximality with respect to text inclusion.
This can be achieved by using a  constraint {\tt cleanup} defined by 
the following rules.
\begin{verbatim}
    vp(_), {!cleanup} <:> true.
    pp(_), {!cleanup} <:> true.
    sentence(_), {!cleanup} <:> true.
    % etc.
    (..., np(_), ... $$ !np(_)), {!cleanup} <:> true.
    cleanup <=> true.
\end{verbatim}
Recall that the exclamation mark combined with the double arrow indicates
simpagation rules: All but those symbols marked by ``{\tt !}'' are removed from
the store.

Assume the following query is issued
\begin{itemize}\item[]
{\tt ?- parse([}{\it string}{\tt]),} {\tt cleanup.}
\end{itemize}
The {\tt cleanup} rules does not affect parsing as there is no {\tt cleanup} constraint
in the store before all {\tt token} constraints have been entered and no
parsing rule can apply anymore.
Now the call to {\tt cleanup} will, via the first set of rules, remove all non-{\tt np} nodes;
these simpagation rules will apply over and over until all such nodes are removed but each application leaves  {\tt cleanup} in the store.
Then the rule concerning {\tt np}s will apply to each occurrence of
one {\tt np} textually included in a larger {\tt np}; recall that
\verb!$$! is the parallel match operator and the three dots are a gap.
The final rule, conveniently written as a CHR rule, will apply when the other
rules are exhausted and thus clean up the {\tt cleanup} constraint.
Left in the constraint store is the set of all maximal {\tt np}s.
\end{example}

\section{Abductive language interpretation in CHRG}\label{abduction-section}
As shown by~\cite{AbdChr2000} and developed
further in~\cite{christiansen-nlulp02}, abduction with integrity constraints
can be implemented in a\break straightforward fashion in CHR, basically by declaring
abducible predicates as constraints: When an abducible atom is called,
it is added to the constraint store and possible integrity constraints will be
triggered automatically.
The approach is limited with respect to negation: Explicit negation of abducibles is
easily implemented by means of an integrity constraint but more general application of
negation-as-failure in background clauses or CHR rules has no obvious representation.

We can illustrate the application to language interpretation in CHRG by means of
an example. Consider the following grammar rule in which $F$ refers to a fact about
the semantical context for a given discourse.
\begin{equation}
     \mbox{\tt a, b, \{$F$\} ::> ab}
     \label{dcg-example}
\end{equation}
If two subphrases referred to by {\tt a} and {\tt b} have been recognized
and the context condition $F$ holds, it is
concluded that an {\tt ab} phrase is feasible, grammatically as well as with
respect to the context.
Language analysis with such rules works quite well when context is completely known
in advance, and a given discourse can be checked to be syntactically and
semantically sound.

Here we provide a solution to the extended problem referred to as {\it language
interpretation}, of finding proper context theory so that an analysis
of an observed discourse is possible.
This involves a transformation of grammar rules as above by moving contextual
predicates to the other side of the implication:
\begin{equation}
     \mbox{\tt a, b ::> \{$F$\}, ab}
     \label{dcg-example-transformed}
\end{equation}
Ituitively it reads: If suitable {\tt a} and {\tt b} are found, it is
feasible to assert $F$ and (thus, under this assumption) to  conclude {\tt
ab}.

Although~(\ref{dcg-example}) and~(\ref{dcg-example-transformed})
are not logically equivalent it is straightforward to formulate and prove correctness
of this transformation as we will see below.

A grammar as depicted by~(\ref{dcg-example}) can be thought of
as part of
a {\it speaker's} capabilities, embedding his knowledge about the context
into language, whereas~(\ref{dcg-example-transformed})
is relevant for a {\it listener} who wants to gain new context knowledge
by an interpretation of the spoken.

\subsection{Abduction as bottom-up deduction}\label{subsection-with-reqr-to-abd-grammars}
The transformation indicated above can be formulated without
detailed assumptions
about the grammar formalism applied,
it may in principle include any kind of transformations,
multiple passes and be based on trees, graphs or something
completely different. The input need not necessarily be strings or sequences but
might also be a combination of sensor signals or multidimensional structures,
e.g., described by means of
Constrained Multiset Grammars~\cite{Marriott94}.

The vocabulary for a language interpretation problem consists of
disjoint sets of constraints referred to as {\it grammar symbols} and
{\it context predicates}.
Grammar symbols are separated into {\it token level} symbols
and {\it phrase level} symbols.

The basic components in a language interpretation scenario are
the following.
\begin{description}
     \item[$\Discourse${\rm:}] A set of ground
     token level atoms giving
     the set of input tokens and their relative order (e.g., sequentially
    or in the shape of a graph for a visual language)
     and, if available, extra information such as prosody, colour, etc.

     \item[$\Context${\rm:}] A set of ground context
     atoms describing a part of the world.

     \item[$\IC${\rm:}] A set  of {\em integrity constraints} which must be satisfied by
$\Context$, each of the form  $H\rightarrow B$ where $H$ is a conjunction of context atoms
and $B$ a conjunction of built-in's and context atoms; however, the total
set of integrity constraints must not be recursive
(or should satisfy some weaker criterion that
guarantees termination).

     \item[$\Phrases${\rm:}] A set of ground phrase level atoms giving
     the phrases contained in the {\it Discourse} that are
     grammatically correct and consistent with {\it  Context}.
     \item[$\Grammar${\rm:}]  A set of formulas for the
     form\[
    \forall(\Constituents\land\Facts\rightarrow\Phrase)\mbox{,}\]
     where $\Constituents$ and $\Phrase$ are nonempty conjunctions of grammar atoms,
     $\Facts$ a conjunction of context atoms.
    Each rule must be {\em range-restricted} in the sense that
    any variable in $\Phrase$ must occur in $\Constituents$ or $\Facts$
    and the grammar must be {\em loop-free} defined analogously
    to definition~\ref{loop-free-def} (for CHRG).
Furthermore, each argument in the lefthand side must be a variable that do
not occur elsewhere in that lefthand side.
\end{description}
We require the following fundamental relation
referred to as {\em faithfulness} between
the components:
\begin{equation}
\cases{
\Grammar\land\Discourse\land\Context\rightarrow\Phrases&\cr
\IC\,\cup\Context\mbox{\rm\ is consistent}&\cr}
\label{l-i-problem}
\end{equation}
This means that the $\Discourse$ and the $\Phrases$
in it are true to the $\Context$ and correctly formulated
with respect to the $\Grammar$.

In case of an ambiguous grammar, we can expect different
interpretations for
different parses of the string.
However, we do not
require the grammar to be unambiguous, but
assume a criterion of {\it unambiguity} of
a set of $\Phrases$ which is particular to the
grammar formalism applied;
a criterion for CHRG is given by definition~\ref{def-unambiguity} above.

Not every pair of unambiguous $\Phrases$ and $\Context$ is interesting:

\begin{definition}
\label{def-competent-interpretation}
     A pair of unambiguous $\Phrases$ and $\Context$ is
     a {\em competent interpretation} of given $\Discourse$
     with respect to given $\Grammar$
     whenever faithfulness and
     the following
     conditions hold:
     \begin{enumerate}
         \item\label{context-minimal}
	(Minimality of $\Context$) If any element is
	removed from $\Context$, faithfulness fails to hold.

         \item\label{phrases-maximal}
	(Maximality of $\Phrases$) If any new element is
	added to $\Phrases$, unambiguity or faithfulness
	fails to hold.

         \item  (Analysis is exhaustive)
	No new elements can be added to $\Context$ which allow
	an extension of $\Phrases$ so that points~\ref{context-minimal}
	and~\ref{phrases-maximal}, and faithfulness are preserved.
     \end{enumerate}
     A {\em language interpretation problem} is a problem,
     given $\Grammar$ and $\Discourse$ of finding a
     competent interpretation.
\end{definition}
The condition of exhaustive interpretation excludes
$\Context=\Phrases=\emptyset$ unless the $\Discourse$ is
completely senseless.

Language interpretation is partly deductive and partly abductive:
The {\it Context} is a premise in~(\ref{l-i-problem}) and by standard 
usage, the
finding of it is an abductive problem.
Identifying phrases is a mainly deductive parsing process,
applying grammar rules over and over,
however, interacting with abduction in order to have
the necessary contextual facts ready.

The translation of a grammar $G$ into an version that can be executed
in a purely deductive way is defined by a transformation $T(G)$ in which
each rule 
\begin{equation}
\forall(C\land F\rightarrow P)
\end{equation}
is replaced by the rule
\begin{equation}
\forall(C \rightarrow \exists\bar z(F\land P))
\end{equation}
where $\bar z$ are the variables
in $F$ that do not occur in $C$.
The fact that $T(G)$ may not be range-restricted indicates some technical
problems that we have to deal with, but it should be emphasized that
$T(G)$ being non-range-restricted does not necessarily indicate that $G$
is too weakly specified: Although a variable in $F$ does not receive a value
by the matching of $C$, it may receive a value later from an integrity constraint
--- or it may remain unbound in case the discourse does not provide
enough information.
The presence of such variables indicates that we cannot expect derivations
to produce ground $\Context$ and $\Phrases$, and an arbitrary grounding
(instantiation of variables) in such cases will produce a more specific
solution than there is evidence for --- even if it is minimal wrt.\ set-inclusion.
This discussion should clarify the following correctness theorems.

\begin{theorem}[Completeness]\label{completeness-of-abductive}
Let $\Grammar$, $T(\Grammar)$ and ground $\Discourse$ be given as above.
If there exist ground $\Context$ and $\Phrases$ so that
faithfulness~(\ref{l-i-problem})
holds with $\Context$ minimal wrt.\ this property, 
then there exist $\Context\,'$ and $\Phrases\,'$ 
so that
\begin{equation}
T(\Grammar)\land\Discourse\land IC\rightarrow \exists(\Context\,'\land\Phrases\,')
        \label{l-i-problem-deductively}
\end{equation}
where $\langle\Context,\Phrases\rangle$ is an instance
of $\langle\Context\,',\Phrases\,'\rangle$.
\end{theorem}

\begin{theorem}[Soundness]\label{soundness-of-abductive}
Let $\Grammar$, $T(\Grammar)$ and ground $\Discourse$ be given as above.
If there exist $\Context\,'$ and $\Phrases\,'$ so that  
\begin{equation}
T(\Grammar)\land\Discourse\land IC\rightarrow\exists(\Context\,'\land\Phrases\,'),
        \label{l-i-problem-deductively-2}
\end{equation}
then there exists a ground instance $\langle\Context,\Phrases\rangle$
of $\langle\Context\,',\Phrases\,'\rangle$
so that $\IC\cup\Context\,'$ is consistent and
\begin{equation}
\Grammar\land\Discourse\land\Context\,'\rightarrow\Phrases\,'
        \label{l-i-problem-2}
\end{equation}
\end{theorem}

\begin{proof}[Proof of theorem~\ref{completeness-of-abductive}]
Let $\Grammar$, $T(\Grammar)$, ground $\Discourse$, $\Context$ and $\Phrases$
be as in the theorem so that~(\ref{l-i-problem}) holds.
Define $G$ to be the set of all ground instances of rules
in $\Grammar$, and let
\[G_0=\{  c\rightarrow p\;|\; (c\land f\rightarrow p)\in G
\,\mbox{\rm and}\, f\in \Context
\}\]
\[G^T=\{  c\rightarrow f\land p\;|\; (c\land f\rightarrow p)\in G
\,\mbox{\rm and}\, f\in \Context\}\]
We have from~(\ref{l-i-problem}) that
\[G\land\Discourse\land\Context\rightarrow\Phrases\]
and from this that
\[G_0\land\Discourse\rightarrow\Phrases\mbox{.}\]
I.e., we have eliminated $\Context$ by using a specialized
grammar. The rules of $G^T$ differs from those of $G_0$ by
introducing on the righthand side an element of $\Context$.
Referring to minimality of $\Context$, we have that
\[G^T\land\Discourse\rightarrow\Context\land\Phrases\mbox{.}\]
Consider now a ``proof'' of $\Context\land\Phrases$ applying a finite sequence
of rules $c_i\rightarrow f_i\land p_i$, $i=1,\ldots,n$ to generate the following sets:
\[C_0= \Discourse,\quad F_0=\emptyset\]
\[C_i=C_{i-1}\cup p_i, \quad c_i\subseteq C_{i-1}\]
\[F_i=F_{i-1}\cup f_i\]
\[C_n=\Phrases,\quad F_n=\Context\]
From this, we construct another parallel proof in which the rules applied are instances
of clauses of $T(\Grammar)$, $(c'_i\rightarrow f'_i\land p'_i)\sigma_i$ where $\sigma_i$ is
a substitution to the variables of $c'_i$ so that
\[C'_0= \Discourse,\quad F'_0=\emptyset\]
\[C'_i=C'_{i-1}\cup p'_i\sigma_i, \quad c'_i\sigma_i\subseteq C'_{i-1}\]
\[F'_i=F'_{i-1}\cup f'_i\sigma_i\]
\[C'_n=\Phrases\,',\quad F'_n=\Context\,'\]
By induction over $i$, it is straightforward to prove that
\[T(\Grammar)\land\Discourse\rightarrow \exists(\Context\,'\land\Phrases\,')\]
and that $\langle\Context,\Phrases\rangle$ is an instance
of $\langle\Context\,',\Phrases\,'\rangle$.
From this,~(\ref{l-i-problem-deductively}) follows immediately.
\end{proof}
\begin{example}
The restriction that
each argument in the head of a grammar rule must a variable that do
not occur elsewhere in that head
is necessary as indicated by the following example.
Let $a/0$, $b/1$, and $c/1$ be grammar symbols, $f/1$ a context predicate
and let $\Grammar$ consist of
\[\mbox{(i)}\;\;\forall x(a\land f(x)\rightarrow b(x))\mbox{,}\quad
\mbox{(ii)}\;\;b(7)\rightarrow c(7)\mbox{.}\]
Then $T(\Grammar)$ consists of (ii) and
\[\mbox{(i$'$)}\;\;\forall x(a\rightarrow f(x)\land b(x))\mbox{.}\]
Given $\Discourse=\{a\}$ and $\Context=\{f(7)\}$ we have that 
$\Phrases=\{a, b(7), c(7)\}$ satisfies the faithfulness condition~\ref{l-i-problem}.
However, a proof using $T(\Grammar)$ will only give
$\Phrases\,'=\{a, \exists x \,b(x)\}$, and it not sound to set this
$x=7$ so that rule (ii) can be applied.
If the head of (ii) had an unrestricted variable instead of a constant,
it
would be possible to relate it to the existentially quantified $\exists x \,b(x)$.
\end{example}

\begin{proof}[Proof of theorem~\ref{soundness-of-abductive}]
Similarly to the proof of theorem~\ref{completeness-of-abductive}.
\end{proof}

\subsection{First version of abduction in CHRG: Locally unambigous grammars}
The general model developed in section~\ref{subsection-with-reqr-to-abd-grammars}
fits perfectly with locally unambigous CHRGs.
For simplicity, we formulate the approach for propagation rule grammars without 
left and right context parts, but it is obvious that it works also in the general case;
especially interesting are CHRGs of simplification rules only that are guaranteed
to be locally unambigous.
(Section~\ref{all-abductive-in-parallel-section} below describes
a generalization to ambiguous grammars.)

Let us define an {\em abductive CHRG} as a grammar with range-restricted rules of
the form
\[\mbox{\it constituents}\mbox{{\tt,\{}\it context-facts}{\tt\}}\prop\mbox{\it nonterminal}\]
in which (cf.\ section~\ref{subsection-with-reqr-to-abd-grammars}) each argument in {\it constituents} and
{\it context-facts} is a unique variable.
The grammar may be extended with a set of
integrity constraints expressed as CHR propagation rules.

Combining theorems~\ref{completeness-of-abductive} and~\ref{soundness-of-abductive}
with the completeness and soundness properties for parsing derivations,
shows that a locally unambiguous,
abductive grammar, written in the format
\[\mbox{\it constituents}\prop\mbox{{\tt\{}\it context-facts}{\tt\}{\tt,}}\mbox{\it nonterminal}\]
produces competent interpretations of the given input string.

The implemented CHRG system does not include this translation but
assumes the user to write abductive grammars directly in the ``translated form'' which
is anyhow the intuitively simplest for someone with experience in CHR
programming.\footnote{The user may, so to speak, use abduction for text interpretation
in this deductive fashion
without being aware that he or she is using a ``nonstandard'' reasoning technique;
abduction works so to speak for free in CHRG.}

\begin{example}
We consider language interpretation of discourses such as
the following.
\begin{equation}
\mbox{\vbox{\hbox{Garfield eats Mickey, Tom eats Jerry, Jerry is mouse,}
             \hbox{Tom is cat, Mickey is mouse.}}}
\label{tom-is-cat-etc}
\end{equation}
What we intend to learn from~(\ref{tom-is-cat-etc}) are
the categories to which the mentioned proper names
belong and which categories that are food items for others.
An interesting question is to which category  Garfield belongs
as this is not mentioned explicitly.
We define the following vocabulary; the
{\tt abducibles} declaration is synonymous with CHR's
{\tt constraints} except that it also introduces predicates for negated
abducibles with integrity constraints that  implement
explicit negation.\footnote{The declaration of an abducible {\tt a}/1
introduces also constraint {\tt a\_}/1 (representing ``not {\tt a}'') and integrity constraint
{\tt a(X), a\_(X) ==> fail}.}
\begin{verbatim}
     abducibles food_for/2, categ_of/2.
     grammar_symbols name/1, verb/1, sentence/1, category/1.
\end{verbatim}
The background theory is the following consisting of integrity 
constraints only.
\begin{verbatim}
     categ_of(N,C1), categ_of(N,C2) ==> C1=C2.
     food_for(C1,C), food_for(C2,C) ==> C1=C2.
\end{verbatim}
I.e., the category for a name is unique, and for the sake of this
example it is assumed that a given category is the food item for at most one
other category.
The following part of the grammar classifies the different tokens.
\begin{verbatim}
     [tom] ::> name(tom).
     ...
     [is]  ::> verb(is).
     ...
     verb(is) -\ [X] <:> category(X).
\end{verbatim}
The last rule applies a syntactic left context part in order to
classify any symbol to the right of an occurrence of ``{\tt is}''
as a {\tt category}.

A sentence such as ``Tom is cat'' is only faithful to a context if
{\tt categ\_of(tom, cat)} holds in it.
So the grammar in the original specification of the current language interpretation
problem may contain the following rule.
\begin{equation}
     \mbox{\vbox{\hbox{\it name($i_1$, $i_2$, N) $\land$ verb($i_2$, 
$i_3$, is) $\land$
     category($i_3$, $i_4$, C) $\land$ categ-of(N,C)}
     \hbox{\it$\rightarrow$  sentence(is(N,C))}}}
\end{equation}
By moving the context condition from the premises to the conclusion
we achieve a rule that can contribute to solve the problem
deductively.
In CHRG it becomes the following:
\begin{verbatim}
     name(N), verb(is), category(C) ::>
        {categ_of(N,C)},
        sentence(is(N,C)).
\end{verbatim}
A sentence such as ``Tom eats Jerry'' is only faithful to a context if
the proper {\tt categ\_of} and {\tt food\_for} facts hold in it.
A CHRG rule with this in its conclusion looks as follows.
\begin{verbatim}
     name(N1), verb(eats), name(N2) ::>
         {categ_of(N1,C1), categ_of(N2,C2), food_for(C1,C2)},
         sentence(eats(N1,N2)).
\end{verbatim}
Let us now trace the processing of the discourse~(\ref{tom-is-cat-etc})
when entered into the constraint store; we record
only the context facts.
``Garfield eats Mickey'' gives rise to
\begin{verbatim}
     categ_of(garfield,X1), categ_of(mickey,X2), food_for(X1,X2).
\end{verbatim}
The ``{\tt X}''s are uninstantiated variables.
The next ``Tom eats Jerry'' gives
\begin{verbatim}
     categ_of(tom,X3), categ_of(jerry,X4), food_for(X3,X4).
\end{verbatim}
``Jerry is mouse'' gives
\verb!categ_of(jerry,mouse)!,
and the background theory immediately unifies {\tt X4} with {\tt mouse}.
In a similar way ``Tom is cat'' gives rise to a unification of {\tt X3}
with {\tt cat} and \verb!food_for(X3,X4)! has become
\begin{verbatim}
     food_for(cat,mouse).
\end{verbatim}
Finally ``Mickey is mouse'' produces \verb!categ_of(mickey,mouse)!
that triggers the first integrity constraint unifying {\tt X2} with
{\tt mouse} and thus the second integrity constraint
sets {\tt X1=cat} and there is no other possibility.
So as part of the solution to this language interpretation problem,
we have found that Garfield is a
cat.
\end{example}
In addition to what we have shown, the
user may also define background theories involving
Prolog rules that include calls to abducibles.
The only restriction is that a call to an abducible must not be embedded in an
application of Prolog's negation by failure.

Interestingly, this form of abduction works also together with a definite clause
grammar: Declare your abducibles as CHRG abducibles (or CHR constraints),
add integrity constraints and apply them in the body of your DCG rules.

\begin{example}\label{dcg-with-abduction-example}
The following DCG together with the declarations of abducibles and
integrity constraints written as CHR rules will produce the same abducibles
as the CHRG described above.
\begin{verbatim}
    name(tom) --> [tom].
    % etc.
    category(mouse) --> [mouse].
    % etc.
    sentence(is(N,C)) -->
        name(N), [is], category(C),
        {categ_of(N,C)}.
    sentence(eats(N1,N2)) -->
        name(N1), [eats], name(N2),
        {categ_of(N1,C1), categ_of(N2,C2), food_for(C1,C2)}.
\end{verbatim}
\end{example}
The DCG+CHR approach to abductive language interpretation works
also correctly for ambiguous grammars as backtracking keeps separated the different
possible parses with their abducibles.

\subsubsection*{Compacting abductive answers}
The final state may include abducible atoms with variables
with the meaning that any ground assignment to such variables (not conflicting
with integrity constraints) represents
a solution to the abductive problem.
Consider as an example the following set of abducible
atoms returned as part of the answer $\{${\tt abd(X)}, {\tt abd(Y)}$\}$.
It may subsume solutions with {\tt X$=$Y} as well as {\tt X$\neq$Y},
e.g., $\{${\tt abd(a)}$\}$, $\{${\tt abd(b)}, {\tt abd(c)}$\}$;
both may be minimal but the application may impose
reasons to prefer the one with fewest elements.

It is possible to extend our method so that it dynamically tries to compact solutions
by equating new abducibles to existing ones as a first choice, and then generate
the other possibilities under backtracking.
In fact, such a step is included in many abduction algorithms,
e.g.,~\cite{kakas-et-al-2000}.

To provide this, we may add for each
abducible predicate, an integrity constraint here shown
for a predicate {\tt abd}
of arity one.
\begin{equation}
\mbox{\tt abd(X), abd(Y) ==> (X=Y ; dif(X,Y))}
\label{fido}
\end{equation}
The semicolon is Prolog's disjunction realized by means of backtracking and
{\tt dif}/2 is a lazy test for syntactic nonidentity that behaves the way we specified
for built-in ``$\neq$'' constraints in section~\ref{CHR-section}.
Whenever a new abducible fact, say {\tt h(a)} or
{\tt h(X)}, is created by the application of some rule,~(\ref{fido})
is applied provided there is another fact {\tt p($t$)}  in
the constraint store.
Notice that~(\ref{fido}) is logically redundant and only affects the execution.

An optimization of~(\ref{fido}) using facilities of the implemented
version of CHR (see~\cite{sicstus-manual} for details) is in place:
\begin{equation}
\mbox{\vbox{\hbox to 1cm{\tt h(X), h(Y)\char'43Id ==> (\char'134+X==Y,
                      unifiable(X,Y)) \char'174\ (X=Y ; dif(X,Y))\hss}
             \hbox{\tt pragma passive(Id)}}}
\label{fido-optimized}
\end{equation}
The {\tt pragma} prevents the rule from being activated
twice due to the symmetry in its head and
the purpose of the guard is to suppress useless applications.

The implemented CHRG system~\cite{CHRGwebsite} includes this compaction
principle as an option.
However, in many cases the problem
does not exist as user-defined integrity constraints may instantiate
and equate abducibles sufficiently during the computation;
this is the case in the example with Garfield and friends above.

\subsection{Evaluation of all  abductive answers in parallel}
\label{all-abductive-in-parallel-section}
The implemented CHRG system incorporates a technique
for keeping track of
the different unambiguous sets of grammar symbols that
are created with a locally ambiguous grammar.
\comment{For the sake of a simple description, we stay to
propagation rules with no context parts but the technique
applies to grammars with arbitrary sorts of rules.
Features of the implemented CHR system~\cite{sicstus-manual}  are applied
that are not covered by the formal definition of
CHR given in section~\ref{CHR-section}.}

Each syntax tree and the abducibles associated with it 
are identified by an index, actually a Prolog variable,
hence referred to in the following as an {\em index variable}.
Grammar symbols (apart from
{\tt token}/1) and
abducibles are given an
extra argument to hold the index.

Whenever a rule applies to syntax nodes with indices
$x_1$,$\ldots$,$x_n$, a new index $x$ is created for the new node.
Fresh copies are made of any abducible with an index among
$x_1$,$\ldots$,$x_n$, but now with $x$ as index.
These constraints are called together with any new abducibles
from the body of the rule (also indexed by $x$).
This activates possible integrity constraints (translated
in a suitable way to cope with indexes; see below).
This may result in a failure and to avoid the whole computation to
stop (as does a failure in a committed choice language such as CHR),
a suitable control structure is embedded in the body of the rule.
If such a failure occurs, the rule simply succeeds but avoids the
creation of a new syntax node (and cleanses the constraint store for the
newly constructed constraints); this effectively stops this branch of computation
but allows other successful syntax trees to continue growing.

The compilation of integrity constraints ensures that
they only apply to abducibles with identical indices.
The compilation of the sample
{\tt father(F1,C) \ttbackslash\ father(F2,C) <=> F1=F2}
shows the principle:
\begin{equation}
\mbox{\tt father(X,F1,C) \ttbackslash\ father(X,F2,C) <=> F1=F2}
\end{equation}
The final state in a derivation contains the collection of all constraints relating
to the different parses; each parse, i.e., each competent interpretation can be printed
out separately.

This implementation principle involves a quite heavy overhead due
to the continual copying of constraints and repeated execution of
integrity constraints that have been executed already.
It is available as an option in the CHRG system.

Obviously this is not an ultimate method for evaluation of all different
abductive interpretations in parallel, but it may give inspiration
for more efficient methods; we discuss this topic in the final section.

\section{Assumption grammars in CHRG}\label{assumption-grammar-section}
As our implementation of abduction has shown, CHRG can work
with different sort of hypotheses passed through the constraint store.
Assumption Grammars~\cite{DahlTarauLi97} (AGs) are similar to abductive grammars
in many respect but differ in that hypotheses are explicitly produced
and explicitly used, possible being consumed.
Assumption grammars provide a collection of operators that makes it possible
to control the scope of these hypotheses which is not possible with
an abductive approach.

We describe here an extension of CHRGs  with a version
of AG which is included in the
available implementation of the system~\cite{CHRGwebsite}.
For simplicity, we describe it in a version that is only correct
for locally unambiguous grammars but it is easily extended to ambiguous grammars
with the technique described for abductive grammars
in section~\ref{all-abductive-in-parallel-section}.

In an AG, the expression {\tt+h(a)} means to assert a linear
hypothesis which can be used once in the subsequent text by means
of the expression {\tt-h(a)} (or
{\tt-h(X)}, binding {\tt X} to {\tt a}) called an {\em expectation}.
Asserting the hypothesis by {\tt *h(a)} means that it can be
used over and over again.
We deviate slightly
from the syntax of~\cite{DahlTarauLi97}
as to achieve a more symmetric notation and
introduce three operators for so-called time-less
hypotheses, \texttt{=+}, \texttt{=-}, and \texttt{=*},
whose meaning are similar except that hypothesis can be
used and consumed in any order.
Compared with the initial proposal for AG, our version
extends also with other features of CHRG, most notably integrity constraints and context
parts.

These operators are defined as constraints in CHR and
can be called from the body of grammar rules.
We introduce the principle by a simplified and incorrect version of
the time-less
versions given by the following CHR rules.
\begin{verbatim}
      =+A, =-B <=>  A=B.
      =*A \ =-B <=> A=B.
\end{verbatim}
\eject
By the first rule, a pair of assumption {\tt=+h(a)} and expectation
{\tt=-h(X)} are removed from the constraint store producing the
effect of binding {\tt X} to {\tt a}.
If assumption {\tt=*h(a)} were used instead, the second
rule can apply to several instances of {\tt=-h($\cdots$)}.
The problems with this implementation are:
\begin{itemize}
      \item  The computation fails in case one of the rules is applied
      for incompatible hypotheses, e.g., {\tt=+h(a)} and {\tt=-g(X)}.

      \item  If two different hypotheses can apply for the same
      expectation {\tt=-h(X)} things go wrong:
      Rule one will only apply one of them and forget
      all about the other one, and rule two applies both
      of them leading obviously to failure.
\end{itemize}
To repair this, we introduce backtracking
and give back hypotheses to the
store when a choice of an expectation-hypothesis pair is given up;
the latter is necessary as CHR uses committed choice.
In order to avoid loops, some book-keeping is added so that
a choice already tested is not tried again.
For {\tt=+} the following is sufficient; the rule for {\tt=*}
is quite similar.
\begin{equation}
\mbox{\parbox{10cm}
{\tt=+A, =-B <=>\\
\hbox to 3em{}(\char92+ has\char95tried\char95rule1(A,B), unifiable(A,B))\\
\hbox to 2em{}|\\
\hbox to 3em{}(A=B ; tried\char95rule1(A,B), =+A, =-B).
}}
\label{splitting-assumptions-rule}
\end{equation}
The predicate \verb!has_tried_rule1! uses CHR facilities to check
whether the indicated instance of the auxiliary constraint
\verb!tried_rule1! is present in the store.
The test for unifiability in the guard is an obvious
optimization which in principle could have been left out.
The operators denoted by prefix {\tt +}, {\tt -}, and {\tt *}
are implemented in a quite similar way, with the CHRG compiler
adding an extra argument corresponding to positions in the input
string; a test that assumption is created textually before expectation
is easily added to the guard.

\begin{example}[Adapted from~\cite{DahlTarauLi97}]\label{example-AG}
We consider sentences with pronouns and coordination
such as ``Martha likes and Mary likes Paul, she hates her''.
We add gender to names and pronouns, and whenever a name appears
as subject or object (in this grammar grouped as {\tt np}s),
an assumption is made that the given name is {\tt acting}.
A pronoun as subject or object gives rise to an expectation for
someone {\tt acting} of appropriate gender.
The principles is shown by the following excerpt.
\begin{verbatim}
      [mary] <:> name(mary, fem).
      [she] <:> pronoun(fem).
      name(X,Gender)  <:> *acting(X,Gender), np(X,Gender).
      pronoun(Gender) <:> -acting(X,Gender), np(X,Gender).
\end{verbatim}
To handle the coordination problem, an incomplete sentence
followed by {\tt and} raises a time-less expectation for
a subject which is met by the assumption produced by the
full sentence at the end.
\begin{verbatim}
  np(A,_), verb(V) /- [and] <:> =-ref_object(B), sentence(s(A,V,B)).
  np(A,_), verb(V), np(B,_) <:> =*ref_object(B), sentence(s(A,V,B)).
\end{verbatim}
One of the possible final states produced for
the sample text above
contains
{\tt sen}\-{\tt ten}\-{\tt ce} symbols with the following attributes:

\medskip\noindent
\hbox to 0pt{}{\tt\ \ \ \ \ \ s(martha,like,paul)},
{\tt s(mary,like,paul)}, and
{\tt s(mary,hate,martha)}.
\end{example}
The AG operators are included in the available CHRG
package~\cite{CHRGwebsite} together with other facilities
of AGs described in~\cite{DahlTarauLi97}.

As mentioned, the CHRG version of AG goes beyond the original proposal by adding integrity
constraints. To see the use of this, consider again example~\ref{example-AG}.
Another final state for the given sentence gives
{\tt s(mary,hate,mary)}.
We can exclude this  by an integrity constraint to prevent that
people hate themselves:
\begin{verbatim}
      sentence(s(A,hate,A)) ::> fail.
\end{verbatim}
In general we can have such rules produce new hypotheses,
e.g., {\tt =*depressed(A)} instead of failing in the rule above,
and combinations of hypotheses can give rise other hypotheses.

\section{Summary and future perspectives}\label{summary-section}
CHR Grammars founded on current constraint logic technology have been
introduced, and their application to aspects of natural language syntax
has been illustrated by small examples.
CHRG can bee seen as a
technologically updated 
ancestor of Definite Clause Grammars: A relative transparent layer
of syntactic sugar over a declarative programming language, providing
both conceivable semantics and fairly efficient implementation.
In CHRG we have replaced Prolog by Constraint Handling Rules.
The result of this shift is a very powerful formalism in which
several linguistic aspects, usually
considered to be complicated or difficult, are included more or less for free:
\begin{itemize}
\item Ambiguity and grammatical errors are hand\-led
in a straightforward way as all different (partial) parses are evaluated
in parallel.
\item Context-sensitive rules,
which are an inherent part of the paradigm,
handle examples of coordination in an immediate way.
\item Abduction, which is 
useful for identifying indirectly implied information,
is expressed directly with no additional computational devices being involved.
\end{itemize}
Context-sensitive rules combined with the ability to handle left-recursion
(as opposed to DCG) are a great help for producing grammars with relatively
few, concise rules without artificial nonterminals; a drawback
is the lack of empty production.

No real-world applications have been developed
in CHRG yet, but we have good expectation for scalability
as selected grammars can run in linear time. Furthermore, the
full flexibility of the underlying CHR and Prolog machinery is available
for optimizations.
Independently, CHRG is available as powerful modeling and prototyping tool.

The approach of using Constraint Handling Rules for language possesses
a potentiality for getting closer to a full integration of lexical, semantic, and pragmatic analysis.
A lexical schism $S$, for example, in the beginning of a discourse may be delayed
until a few sentences later when the semantic context is identified so that
$S$ can be resolved and, thus, that analysis can resume
for the first sentence.

Although being a very powerful system in itself, CHRG and
the examples we have tested appear only to touch upon 
the surface of what is possible. It is obvious that weights
can be added and used to suppress all but the most likely interpretation,
and arbitrary constraint solvers can be incorporated in this process.
Although presented here as a strict bottom-up paradigm, it is possible to
add top-down guidance to parsing in CHR and CHRG which is useful in order
to prevent local ambiguity to result in the creation of a lot of useless constraints;
top-down guidance is applied in the work of~\cite{ChristiansenDahl2002,ChristiansenDahl2003}
but for other purposes.

The basic principle may seem quite na\"\i ve, almost too na\"\i ve, just applying
grammar rules bottom-up over and over until the process stops.
However, we can rely now on the underlying, well-established computational
paradigm of CHR for such rules-based
computations. Furthermore, the approach can profit
from any future improvements of CHR and similar
deductive systems.

As noticed above, our implementation in CHR for parallel evaluation of different
abductive interpretations of a discourse is far from ideal, but it may serve
as an important source of inspiration for
the development of better systems.
Instead of simulating several constraint stores by means of extra index arguments,
it seems obvious to apply a sort of shared representation for the different
stores so that copying of constraints is avoided.

\section*{Acknowledgements}
Part of this work has been carried out while the author
visited Simon Fraser University, Canada, partly supported by the
Danish Natural Science Council;
thanks to Ver\'onica Dahl for helpful discussion and providing a stimulating environment.
This research is supported in part by the OntoQuery
funded by the Danish Research Councils, and
the IT-University of Copenhagen.

\end{document}